\documentclass[runningheads]{styles/llncs}

\usepackage{styles/eccv}

\usepackage{styles/eccvabbrv}
\usepackage{booktabs}
\usepackage{multirow}
\usepackage{graphicx}
\usepackage{tabularx}
\usepackage{lmodern}
\usepackage{hyperref}
\usepackage{styles/orcidlink}
\graphicspath{{figures/}}
\hypersetup{hidelinks}

\begin{document}

\title{Reliability-Aware 3D Geometric Injection for Universal Person Re-identification}

\titlerunning{Universal Monocular 3D-Enhanced ReID}

\author{Bohan Su\inst{1}\textsuperscript{*}\orcidlink{0009-0006-9405-3817} \and
Jiashuo Wang\inst{1}\textsuperscript{*}\orcidlink{0009-0007-0073-7677} \and
Fangyi Liu\inst{1}\textsuperscript{\ensuremath{\dagger}}\orcidlink{0000-0001-8815-0254} \and
Mang Ye\inst{1}\textsuperscript{\ensuremath{\dagger}}\orcidlink{0000-0003-3989-7655}}
\authorrunning{B.~Su et al.}
\institute{\textsuperscript{1} National Engineering Research Center for Multimedia Software,\\
School of Computer Science, Wuhan University, Wuhan, China\\
\email{\{bohansu,jiashuowang,fangyiliu,yemang\}@whu.edu.cn}}

\maketitle
\begingroup
\renewcommand{\thefootnote}{\fnsymbol{footnote}}
\footnotetext[1]{Equal contribution. \textsuperscript{\ensuremath{\dagger}} Co-corresponding authors.}
\endgroup

\begin{abstract}
Universal person re-identification (ReID) aims to retrieve pedestrian identities across diverse real-world scenarios, including severe occlusions, clothing changes, and cross-modality shifts, within a unified model.
However, existing 2D representations fundamentally struggle with spatial ambiguities due to a lack of depth and topological awareness, while naively introducing monocular 3D priors often causes severe negative transfer due to geometric estimation noise under extreme visual degradation.
To safely harness the clothing-invariant and canonical structural properties of 3D geometry, we propose UniGeo, a Universal Monocular 3D-Enhanced ReID framework driven by a Consistency-Aware Reliability Gate and Dual-Stream Residual Fusion.
Specifically, the processing of 3D information is strategically decoupled into geometric extraction and dynamic utilization.
To provide pure structural compensation, we project monocular 3D parameters into kinematic joint representations, explicitly capturing instance-level geometric topology to resolve appearance-based ambiguities.
To robustly incorporate these cues without perturbing the reliable 2D feature space, we isolate the 3D prior as a late-stage structural residual; modulated by the consistency-aware gate, this mechanism adaptively filters geometric noise and enables controlled fallback to the pure 2D baseline.
Extensive experiments show that our method improves challenging, structure-sensitive scenarios while preserving competitive performance on clean domains. Code is available at \url{https://github.com/BohanSu/UniGeo}.
\end{abstract}

\section{Introduction}

Person re-identification (ReID) aims to retrieve the same individual across non-overlapping camera views, serving as a fundamental capability for large-scale video surveillance systems and long-term identity tracking applications \cite{zheng2016person, ye2021deep}. While recent deep learning baselines have achieved remarkable progress in closed-set scenarios with controlled conditions \cite{luo2019bag, he2021transreid}, real-world deployments demand robust identification under severe and unpredictable appearance corruptions. These corruptions span a wide spectrum of challenges, including partial occlusions caused by obstacles or crowds \cite{miao2019pose, zheng2019pose}, dramatic clothing changes across sessions \cite{yang2019person, gu2022clothes}, and cross-sensor variations in illumination, resolution, and even imaging modality (\eg visible-to-infrared transitions) \cite{wu2017rgb, ye2021deep}.

This practical imperative motivates the paradigm of \emph{Universal ReID} \cite{zheng2024versatile, chen2023towards}: instead of maintaining separate specialist models for each scenario, the goal is one model that generalizes across heterogeneous domains. Maintaining parallel expert models is expensive, architecturally brittle, and difficult to scale in practice. These domains include standard holistic benchmarks such as Market-1501 \cite{zheng2015scalable} and MSMT17 \cite{wei2018person}, clothing-change scenarios such as PRCC \cite{yang2019person}, severe occlusion benchmarks such as Occluded-Duke \cite{miao2019pose}, and cross-modality datasets such as SYSU-MM01 \cite{wu2017rgb}. The promise of Universal ReID is to reduce task-specific silos and support seamless deployment across varied real-world conditions. Yet universality remains difficult because these domains differ in both low-level visual statistics (viewpoint, style, resolution) and semantic structures (pose articulation, body topology, and modality characteristics) \cite{wang2018person, choi2020hi}.

\begin{figure}[t]
  \centering
  \includegraphics[width=\linewidth]{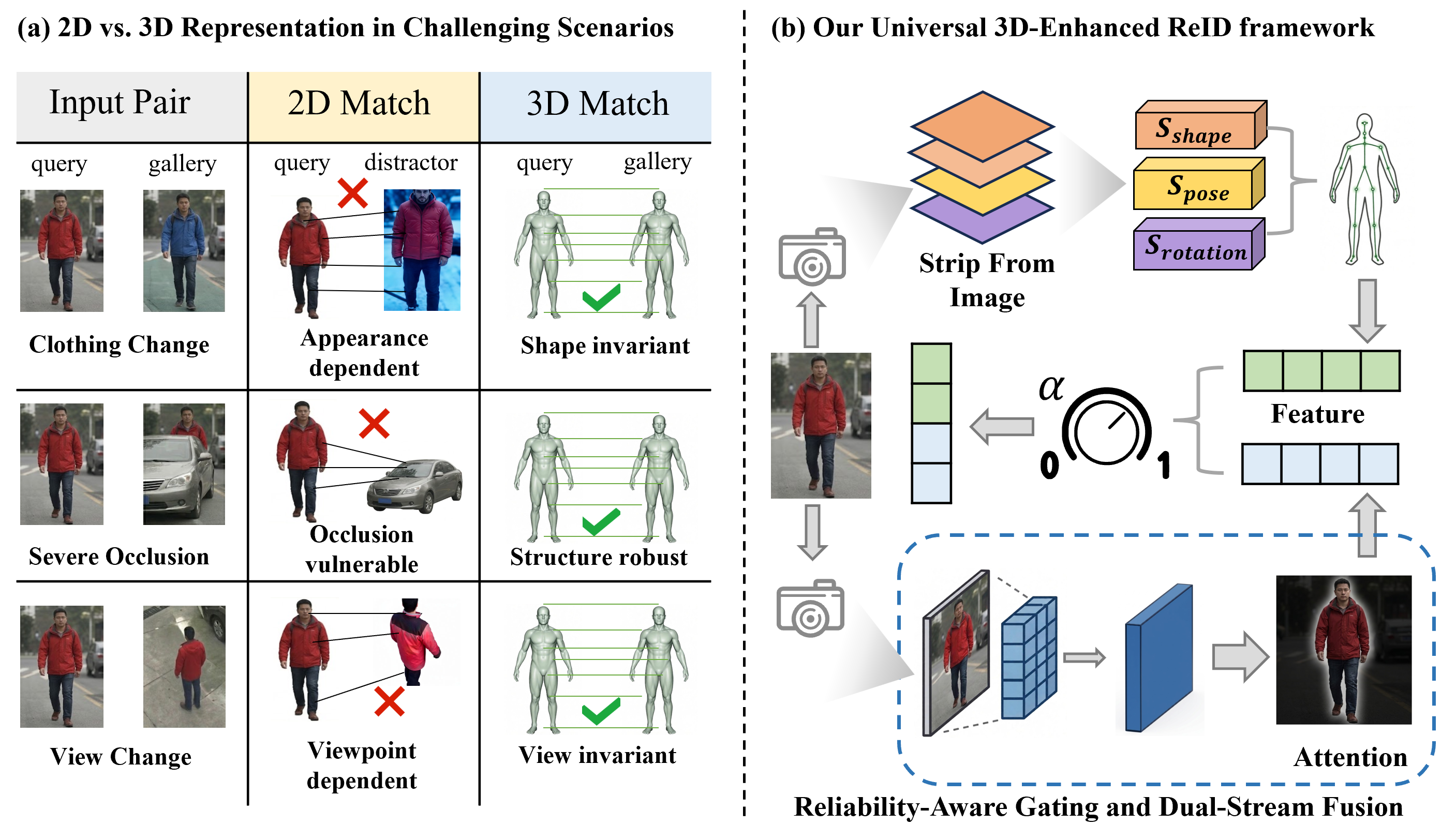}
  \caption{(a) Standard 2D pipelines are vulnerable to severe appearance corruptions (\eg heavy occlusion, clothing changes). (b) Our 3D-aided pipeline leverages kinematic-aware geometry and reliability-aware gating to conditionally fuse 3D structural cues with 2D features, ensuring robust identity representation.}
  \label{fig:intro}
\end{figure}

Recent 2D advances include transformer-based representation learning \cite{he2021transreid, zhu2023aaformer}, vision-language transfer \cite{li2023clip}, and prompt-driven universal adaptation \cite{zheng2024versatile}. Together, these methods provide increasingly strong visual foundations for cross-domain generalization. Yet a critical limitation persists: their identity evidence still relies mainly on 2D appearance cues from textures and color patterns \cite{geirhos2018imagenet}. When such cues fail under heavy occlusion \cite{yan2021occluded}, extreme lighting variation \cite{wu2017rgb}, or clothing change \cite{hong2021fine}, these representations suffer spatial ambiguities and semantic collapse. Consequently, as illustrated in \cref{fig:intro}(a), current universal pipelines still face a structural bottleneck when appearance information is heavily corrupted or uninformative \cite{chen2021learning}.

Introducing 3D geometry is therefore a principled way to overcome this bottleneck \cite{chen2021learning, zhu2020identity}. Recent advances in monocular 3D human reconstruction, especially transformer-based models such as 4DHumans \cite{goel2023humans}, have made geometry extraction practical for ReID. We adopt 4DHumans as our default SMPL extractor following recent practice, while keeping the framework model-agnostic: any SMPL-based estimator can provide the upstream geometry. The core challenge is not choosing a 3D estimator, but using its outputs safely. Monocular 3D reconstruction remains inherently ill-posed \cite{zhang2025survey}, and even state-of-the-art estimators become unreliable under severe visual degradations (blur, truncation, heavy occlusion) or domain shifts (\eg RGB-to-infrared transitions in cross-modality scenarios). Naive static fusion that unconditionally injects such noisy estimates can propagate geometric errors into the feature space, ultimately degrading rather than improving ReID discrimination performance \cite{somers2024keypoint}.

To overcome these limitations, as depicted in \cref{fig:intro}(b), our main contribution is to treat 3D geometric cues as \emph{conditional} structural evidence rather than a uniformly trusted auxiliary input \cite{kendall2017uncertainties}. We instantiate this paradigm through a two-part architecture: (1) a kinematic-aware geometry extraction branch that projects off-the-shelf SMPL parameters \cite{loper2015smpl} into structured pose representations $\mathbf{f}_{pose}$ capturing body topology, and (2) a reliability-aware gating mechanism that dynamically estimates geometric reliability by evaluating cross-modal consistency \cite{ramachandram2017deep}. This gating mechanism provides robustness even in demanding cases, such as cross-modality inputs where RGB-trained 3D estimators can become unreliable. It achieves this by detecting visual-geometric discrepancies and safely suppressing unreliable structural signals, thereby preventing negative transfer while preserving strong 2D baselines.

In summary, our key contributions are threefold:
\begin{itemize}
    \item We reformulate monocular 3D geometry as \emph{conditional structural evidence} for Universal ReID, rather than as a replacement for 2D appearance. This perspective is designed for cases where appearance cues become ambiguous under occlusion, clothing change, or modality shift.
    \item We propose UniGeo, which extracts a kinematic-aware structural representation from off-the-shelf SMPL estimates \cite{goel2023humans, loper2015smpl} and injects it through a consistency-aware gate. The model activates geometry when it is helpful and suppresses unreliable 3D when visual-geometric mismatch indicates high risk of negative transfer.
    \item UniGeo delivers clear improvements where structural cues matter most, including +3.0\% Rank-1 on PRCC, +0.9\% on Occluded-Duke, and +1.2\% on SYSU-MM01, while remaining competitive on standard benchmarks such as Market-1501, MSMT17, and CUHK03. These results confirm that reliability-aware geometric injection strengthens robustness under domain corruption without sacrificing standard-domain performance.
\end{itemize}

\section{Related Work}
\label{sec:related_work}

\subsubsection{From Specific to Universal ReID.}
Traditional ReID architectures yield strong performance under closed-world assumptions \cite{luo2019bag, ye2021deep, he2021transreid, li2023clip}. To address real-world domain shifts, existing works mainly target isolated challenges: part-aware and pose-guided alignments for severe occlusion \cite{miao2019pose, li2021diverse, zhu2020identity, gao2020pose, yan2021occluded, wang2022feature, somers2023body}; shape-aligned and gait-assisted disentanglement for clothing changes \cite{yang2019person, hong2021fine, gu2022clothes, jin2022cloth, chen2023beyond, pang2025identity, liu2023dual, liu2024cloth}; and modality compensation for discrepancies \cite{wu2017rgb, ye2021deep, feng2023shape, ren2024implicit}.

While these specialized models excel individually, deploying parallel expert models for unpredictable visual corruptions is impractical. Recent universal frameworks, such as VersReID \cite{zheng2024versatile} and UNIReID \cite{chen2023towards}, unify diverse domains by routing features through scene-specific prompts or task-aware learning, while recent cloth-generalized studies further emphasize robustness under severe appearance shifts \cite{liu2023dual, liu2024cloth}. However, these approaches are still primarily optimized around appearance-driven domain adaptation. As a complementary perspective, we advance the universal paradigm by exploring how 3D auxiliary information can resolve microscopic instance-level structural distortions when texture cues fail. This perspective is particularly relevant when body layout and pose geometry remain semantically informative even after clothing textures or modality-specific color statistics become unreliable. By dynamically injecting monocular 3D priors through a reliability-aware gating mechanism, our dual-stream framework provides robust structural compensation across deployments.

\subsubsection{Monocular 3D Human Reconstruction.}
Recovering 3D human parametric models (\eg SMPL \cite{loper2015smpl, bogo2016keep}) from monocular images provides a solid foundation for extracting structural priors robust to 2D pixel corruption. The field has rapidly evolved from early end-to-end regression frameworks \cite{kanazawa2018end, omran2018neural} to methods exploiting pixel-aligned supervision \cite{kolotouros2019learning, zhang2021pymaf, moon2020i2l} and spatial robustness mechanisms \cite{kocabas2021pare, li2022cliff, jiang2020coherent, zhu2024dpmesh} for handling complex articulations. Recently, Transformer-based architectures \cite{lin2021end, lin2021mesh, cho2022cross, lin2023one} have predominated, culminating in state-of-the-art foundation models like 4DHumans \cite{goel2023humans}.

Despite this progress, in-the-wild monocular 3D recovery remains fundamentally ill-posed. As highlighted by probabilistic models \cite{kolotouros2021probabilistic} and recent surveys \cite{zhang2025survey}, 2D projection introduces depth ambiguity, which is further amplified by severe occlusions and complex scenes. As a result, geometric predictions often become unstable near object boundaries and occluded regions. This intrinsic uncertainty introduces estimation noise, so 3D outputs must be integrated carefully when serving downstream recognition tasks.

\subsubsection{3D-Assisted Person Re-identification.}
Historically, 3D information was introduced for video sequence aggregation \cite{wu20193}, later expanding to novel modalities like WiFi \cite{ren2023person}. In RGB/IR domains, early works leverage 3D models for data augmentation \cite{wang2020surpassing, sun2019dissecting, zhang2021unrealperson, zheng2022parameter}. To tackle clothing changes, while some baselines mine invariants from RGB modalities only \cite{gu2022clothes}, 3D-assisted methods leverage meshes, multi-view transformations \cite{yu2024mv}, and skeleton dynamics \cite{rao2021sm, rao2023transg, joseph2025clothes} for shape-aware alignment \cite{chen2021learning, miao2019pose, somers2024keypoint, ren2024implicit}. Recent advances further establish 2D-3D correspondences \cite{wang2023exploring} or learn clothing-invariant 3D representations \cite{liu2023learning}.

While robust against specific texture variations, these pioneering works assume reliable geometric estimation. However, under unpredictable visual corruptions in Universal ReID, off-the-shelf estimators can produce distorted structures, and unconditionally injecting them can induce negative transfer. Advancing a noise-tolerant philosophy, our approach differs from static or dynamic skeleton encodings \cite{rao2021sm, rao2023transg, joseph2025clothes} and deterministic 3D fusions \cite{liu2023learning}. Instead of rigid 2D-3D correspondences \cite{wang2023exploring} or pure RGB mining \cite{gu2022clothes}, our Reliability-Aware Geometric Injection utilizes kinematic decoupling and a consistency-aware gating mechanism to dynamically modulate 3D fusion. This selectively harnesses geometric invariants when reliable, while limiting estimation noise during extreme spatial-texture conflicts and improving robustness in cross-modality settings \cite{ren2023person}.

\section{Method}
\label{sec:method}

\begin{figure}[t]
  \centering
  \includegraphics[width=\linewidth]{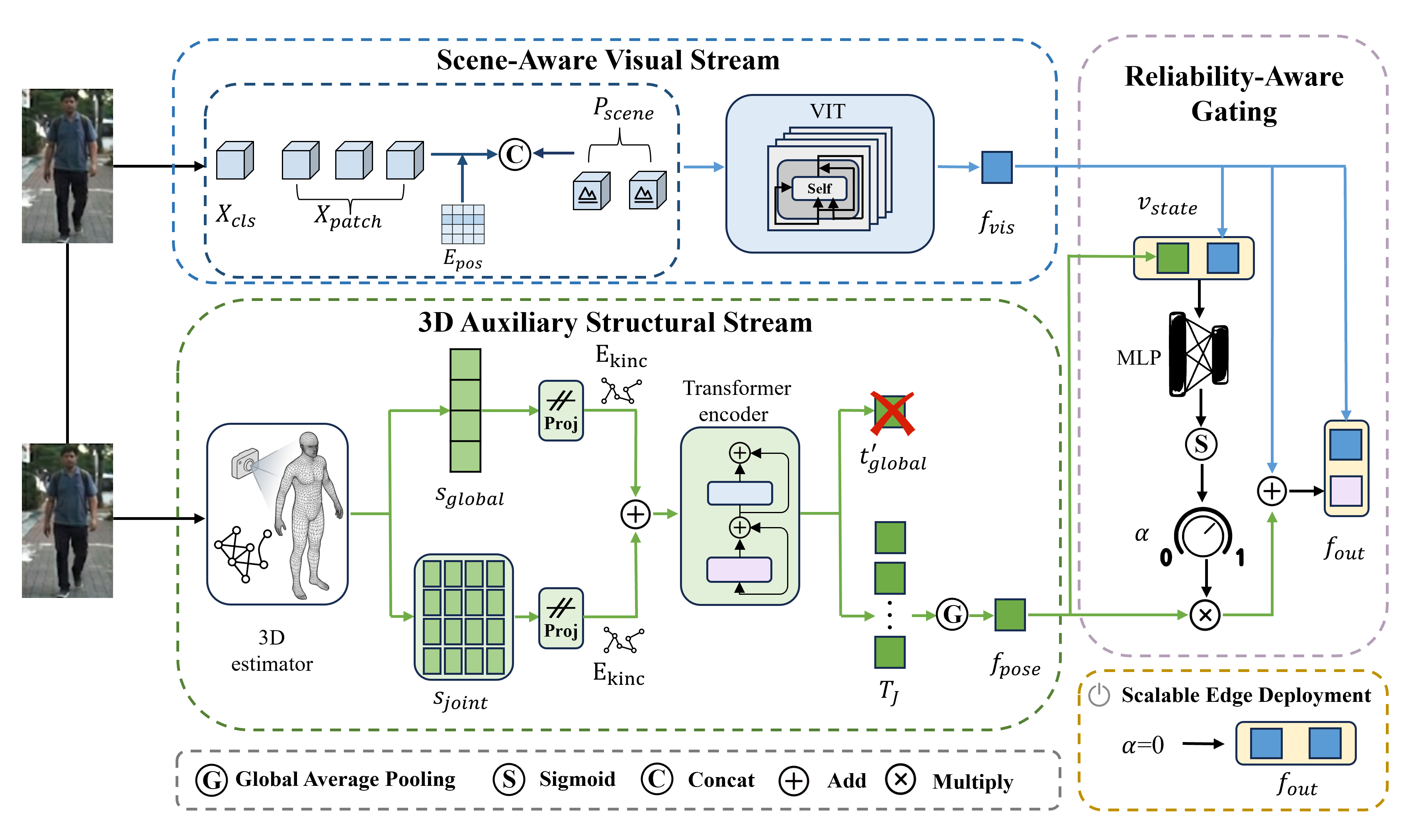}
  \caption{Overview of our UniGeo framework. A Scene-Aware Visual Stream (\cref{subsec:visual_backbone}) extracts 2D textures $\mathbf{f}_{vis}$, while an Auxiliary Structural Stream (\cref{subsec:pose_encoder}) models decoupled 3D geometric features $\mathbf{f}_{pose}$. Guided by cross-modal discrepancy, a Reliability-Aware Gate predicts scalar $\alpha$ to modulate the dual-stream residual fusion (\cref{subsec:fusion}) under end-to-end optimization (\cref{subsec:optimization}).}
  \label{fig:framework}
\end{figure}

The overall pipeline of our proposed UniGeo framework is illustrated in \cref{fig:framework}. With the core objective of extracting robust representations in unpredictable environments, our architecture strategically decouples feature learning into two synergistic streams, safely integrating 3D geometry as structural compensation.

In this section, we first introduce the Scene-Aware Visual Stream (\cref{subsec:visual_backbone}), which provides a 2D baseline for resolving macroscopic domain shifts. We then detail the 3D Auxiliary Structural Stream (\cref{subsec:pose_encoder}), which extracts kinematic-aware 3D topological features to mitigate instance-level corruptions such as severe occlusions. Next, given the inherent instability of monocular 3D estimators in the wild \cite{zhang2025survey, kocabas2021pare}, we present a Reliability-Aware Geometric Gate and a Dual-Stream Fusion strategy (\cref{subsec:fusion}) to prevent geometric noise propagation. Finally, we describe the end-to-end optimization scheme and cached-geometry deployment and inference strategy (\cref{subsec:optimization}).

\subsection{Scene-Aware Visual Representation}
\label{subsec:visual_backbone}

In real-world scenarios characterized by macroscopic data distribution discrepancies (\eg illumination shifts), conventional 2D backbones often struggle to disentangle identity features from environmental noise. Inspired by prompt tuning, we establish a Scene-Aware Visual Stream based on a Vision Transformer (ViT) \cite{dosovitskiy2020image, zheng2024versatile} to explicitly incorporate environmental contexts, thereby extracting domain-resilient appearance textures.

Formally, an input image $\mathbf{x} \in \mathbb{R}^{H \times W \times 3}$ is partitioned and linearly projected into patch embeddings $\mathbf{X}_{patch} \in \mathbb{R}^{N \times D}$, where $D$ is the embedding dimension. To achieve scene-awareness, we introduce learnable scene-specific prompts $\mathbf{P}_{scene} \in \mathbb{R}^{K \times D}$ ($K$ denotes the prompt length). Following the protocol in VersReID~\cite{zheng2024versatile}, these prompts are conditioned on dataset-level scene labels rather than specific camera IDs or raw environmental tags; specifically, each dataset is assigned a scene label denoting its primary ReID challenge (\eg general, low-resolution, occlusion). During multi-dataset training, these labels guide scene-prompt selection to encode domain knowledge. Prepended with a class token $\mathbf{x}_{cls} \in \mathbb{R}^{1 \times D}$, the patch tokens are first supplemented with spatial position embeddings $\mathbf{E}_{pos} \in \mathbb{R}^{(1 + N) \times D}$. The scene prompts are then appended to the sequence end:
\begin{equation}
    \mathbf{Z}_0 = \big[ [\mathbf{x}_{cls}; \mathbf{X}_{patch}] + \mathbf{E}_{pos}; \; \mathbf{P}_{scene} \big].
\end{equation}
After processing through $L$ Transformer layers, the state of the class token is extracted as the global visual representation $\mathbf{f}_{vis} \in \mathbb{R}^{D}$.

While effective against macroscopic domain shifts, this 2D representation inherently relies on local textures and appearance cues \cite{geirhos2018imagenet}. Consequently, under severe instance-level structural corruptions (\eg heavy occlusions), it inevitably suffers from spatial ambiguities due to the lack of topological awareness, which naturally motivates our 3D structural branch.

\subsection{Geometric Feature Extraction}
\label{subsec:pose_encoder}

As analyzed above, 2D visual representations inevitably collapse under severe instance-level corruptions, demanding robust structural compensation. To this end, we introduce a lightweight 3D structural branch utilizing an off-the-shelf monocular 3D estimator, 4DHumans \cite{goel2023humans}, to extract the SMPL \cite{loper2015smpl} parameter vector $\mathbf{s} \in \mathbb{R}^{82}$. Mathematically, this parameter vector encapsulates highly heterogeneous physical quantities, including global spatial orientation and local joint articulations. Treating it merely as a homogeneous black-box vector processed by a naive multi-layer perceptron (MLP) would overlook the explicit physical semantics and the kinematic dependencies among body parts. Motivated by this, we design a kinematic-aware Pose Encoder ($E_{pose}$) to explicitly decouple these components and model their structural topology.

Specifically, to preserve physical semantics, we decouple the 82-dimensional SMPL vector into two components: global parameters $\mathbf{s}_{global} \in \mathbb{R}^{13}$, which encode 3D global rotation and 10-D shape proportions, and local joint parameters $\mathbf{s}_{joint} \in \mathbb{R}^{23 \times 3}$, which represent the relative 3D rotations of 23 body joints.

Instead of mixing these disparate spaces prematurely, two independent projection pathways (linear layers followed by GELU) map them into the latent dimension $D$. This yields a global token $\mathbf{t}_{global} \in \mathbb{R}^{1 \times D}$ and a sequence of local joint tokens $\mathbf{T}_{local} \in \mathbb{R}^{23 \times D}$. To capture kinematic topology, we concatenate these tokens, add positional embeddings $\mathbf{E}_{kine} \in \mathbb{R}^{(1+23) \times D}$, and process the sequence with a lightweight Transformer encoder:
\begin{equation}
    [\mathbf{t}'_{global}; \mathbf{T}'_{local}] = \text{Transformer}\big([\mathbf{t}_{global}; \mathbf{T}_{local}] + \mathbf{E}_{kine}\big).
\end{equation}

Global camera shifts and absolute spatial coordinates often introduce severe view-dependent interference in wild environments. To reduce this variance and focus on articulated human posture, we adopt a two-stage design: during Transformer processing, the global parameters $\mathbf{s}_{global}$ (containing identity-relevant shape factors such as limb proportions and body structure) are fed as input to enrich the joint tokens through cross-attention; however, for the final output, we discard the updated global token $\mathbf{t}'_{global}$ and retain only the 23 refined joint tokens, denoted as $\mathbf{T}_J \in \mathbb{R}^{J \times D}$ (where $J=23$), to improve view invariance. This configuration—using both $\mathbf{s}_{global}$ and $\mathbf{s}_{joint}$ as Transformer input while pooling $\mathbf{f}_{pose}$ exclusively from joint tokens—is validated as the optimal topology in our ablation study (\cref{tab:ablation_modality}). Subsequently, a global average pooling operation is applied across the joint token sequence to distill a compact geometric feature:
\begin{equation}
    \mathbf{f}_{pose} = \frac{1}{J} \sum_{j=1}^{J} \mathbf{t}_{J,j},
\end{equation}
where $\mathbf{t}_{J,j} \in \mathbb{R}^{D}$ denotes the $j$-th joint token within the sequence $\mathbf{T}_J$.
Independent of fragile 2D textures, $\mathbf{f}_{pose}$ provides an explicit structural anchor, laying a solid foundation for the subsequent reliability-aware fusion.

\subsection{Reliability-Aware Gating and Dual-Stream Fusion}
\label{subsec:fusion}

While the aforementioned 3D prior offers valuable topological compensation, monocular 3D reconstruction is fundamentally an ill-posed problem. Under severe visual degradations (\eg heavy truncations), the estimator inevitably produces erroneous geometries. Blindly fusing such geometric noise would corrupt the identity representation, leading to severe negative transfer. To systematically prevent this, we hypothesize that the reliability of the 3D estimation is highly correlated with its spatial consistency with the 2D visual semantics. Consequently, we propose a Reliability-Aware Geometric Gate coupled with a Dual-Stream Residual Fusion strategy to achieve risk-aware structural integration.

\textbf{Consistency-Aware Reliability Gate.} To explicitly evaluate the cross-modal discrepancy, we concatenate the global visual representation and the geometric feature to form a joint state vector $\mathbf{v}_{state} = [\mathbf{f}_{vis}; \mathbf{f}_{pose}] \in \mathbb{R}^{2D}$. Rather than relying on heuristic distance metrics, $\mathbf{v}_{state}$ is fed into a lightweight multi-layer perceptron (MLP) to implicitly model the geometry-visual alignment and predict a consistency-aware reliability coefficient $\alpha$:
\begin{equation}
    \alpha = \sigma\left(\text{MLP}(\mathbf{v}_{state})\right) \in (0, 1),
\end{equation}
where $\sigma(\cdot)$ denotes the Sigmoid function. To maintain computational efficiency, the MLP employs a bottleneck architecture, projecting an input of dimension $2D$ into a compact hidden space of dimension $D/4$ before mapping it to the final scalar. Driven by end-to-end optimization, this gate acts as an adaptive reliability filter, suppressing $\alpha$ when unresolvable divergence occurs between the 2D texture and the estimated 3D geometry.

\textbf{Dual-Stream Residual Fusion.} To safely integrate the geometric prior without compromising the stability of the 2D baseline, we isolate the 3D structural intervention as a controlled residual. The final ReID representation $\mathbf{f}_{out} \in \mathbb{R}^{2D}$ is formulated as a concatenation-residual hybrid:
\begin{equation}
    \mathbf{f}_{out} = \big[ \mathbf{f}_{vis}; \; \mathbf{f}_{vis} + \alpha \cdot \mathbf{f}_{pose} \big].
\end{equation}
This late-stage formulation preserves the stability of the visual stream. The first half of $\mathbf{f}_{out}$ preserves the pure 2D visual feature semantics as an uncorrupted anchor. The second half acts as a dynamically modulated structural residual. When severe visual degradation triggers $\alpha \approx 0$, the structural residual reduces to $\mathbf{f}_{vis}$, completely halting geometric noise propagation. This design lets the dual-stream representation revert to the pure 2D feature space, providing a stable fallback against 3D estimation failures.

\subsection{Optimization and Scalable Edge Deployment}
\label{subsec:optimization}

To rigorously validate that our performance improvements stem from the proposed geometric gating mechanism rather than auxiliary training tricks, we adhere to a standard ReID optimization paradigm. The dual-stream framework is trained end-to-end, with the overall objective imposed exclusively on the final fused representation $\mathbf{f}_{out}$:
\begin{equation}
    \mathcal{L}_{total} = \mathcal{L}_{cls}(\mathbf{p}, y) + \lambda \cdot \mathcal{L}_{tri}(\mathbf{f}_{out}, y),
\end{equation}
where $\mathbf{p}$ denotes the identity logits from a linear classifier, $\mathcal{L}_{cls}$ is cross-entropy with label smoothing, $\mathcal{L}_{tri}$ is the batch-hard triplet loss, $y$ is the identity label, and $\lambda$ weights the triplet term.

\textbf{Cached Geometry and Fallback.} Executing monocular 3D estimators on resource-constrained edge devices is often computationally prohibitive. Our late-fusion architecture addresses this through offline-cached geometry extraction and an inference-time fallback mechanism.\footnote{This fallback removes online 3D extraction at inference, but not all overhead: concatenation still doubles the descriptor dimensionality from $D$ to $2D$, and geometry extraction remains an offline preprocessing step.} By manually enforcing $\alpha=0$, the structural stream is bypassed, degenerating the final representation to $\mathbf{f}_{out}=[\mathbf{f}_{vis};\mathbf{f}_{vis}]$. Since the cosine similarity between such duplicated vectors is mathematically equivalent to their pure 2D counterparts ($\mathbf{f}_{vis}$), the retrieval behavior remains compatible with the 2D baseline without feature distribution shifts. In practice, this design decouples accuracy-oriented geometric preprocessing from lightweight deployment constraints. When cached geometry is available, the model can exploit structural cues; when it is not, the same retrieval pipeline still operates in a visual-only mode without changing the similarity rule.

\section{Experiments}
\subsection{Experimental Setup}
\textbf{Datasets.} We evaluate on 9 benchmarks grouped into 5 scenarios: \textbf{A. Standard Holistic} (Market-1501 \cite{zheng2015scalable}, MSMT17 \cite{wei2018person}, and CUHK03 \cite{li2014deepreid} using the 1367/100 labeled split), \textbf{B. Clothing-Change} (PRCC \cite{yang2019person}, Celeb-ReID \cite{huang2019celebrities}), \textbf{C. Occlusion} (Occluded-Duke \cite{miao2019pose}), \textbf{D. Cross-Modality} (SYSU-MM01 \cite{wu2017rgb}), and \textbf{E. UAV \& Wild} (UAV-Human \cite{li2021uav}, AG-ReID.v2 \cite{nguyen2024ag}). We report Rank-1 and mAP without re-ranking.

\textbf{Implementation Details.}
All models use ViT-Base ($384 \times 128$) with ImageNet pre-training, trained on a single RTX 3090 GPU throughout all experiments reported in this paper. We employ universal joint training with domain-balanced sampling for 180 epochs using SGD (momentum 0.9, lr=$4 \times 10^{-4}$, weight decay=$1 \times 10^{-4}$, cosine annealing), batch size 120 (30 IDs $\times$ 4 images). The loss combines cross-entropy and triplet loss (margin 0.3) with equal weights. SMPL parameters are extracted offline using the pre-trained 4DHumans~\cite{goel2023humans} model (frozen weights, 25 FPS on RTX 3090, $\sim$40ms per image) and cached to disk, eliminating online 3D estimation overhead during training and inference. The 3D stream uses a 2-layer Transformer encoder (hidden dim 768, 8 attention heads). The gating MLP has architecture [1536→768→1] with GELU activation. For data augmentation alignment, random horizontal flips are synchronized by mirroring SMPL joint x-coordinates; random cropping (scale 0.75-1.0) and random erasing (p=0.5) apply only to the visual stream, forming an asymmetric augmentation strategy that exposes the gate to spatially misaligned cases during training and improves robustness at inference.

\textbf{Baseline Configuration.} Our pure 2D baseline adopts the VersReID~\cite{zheng2024versatile} architecture with scene-aware prompts, extended to include UAV-Human and AG-ReID.v2 datasets for comprehensive universal evaluation. Both baseline and our full model are trained under identical protocols (180 epochs, CUDA 12 environment) to ensure fair comparison. This controlled setup isolates the contribution of our reliability-gated 3D injection mechanism.
Unless otherwise stated, we report raw results on nine representative benchmarks and transfer protocols in Tables~\ref{tab:holistic}--\ref{tab:uav_ag_results}. The 12-protocol aggregate in \cref{tab:fusion_evolution} further includes all four AG-ReID.v2 transfer settings together with UAV-Human.

\subsection{Comparison with State-of-the-Art Methods}
\begin{table}[!tb]
\centering
\caption{\textbf{Comparison with other methods on standard holistic ReID benchmarks (Market-1501, MSMT17, and CUHK03).} Each table uses a unified method column, and ``-'' denotes results not reported in the original papers. ``*'' denotes using an overlapping patch embedding layer.}
\label{tab:holistic}
\begingroup
\setlength{\tabcolsep}{3.5pt}
\renewcommand{\arraystretch}{1.03}
\scriptsize
\begin{tabular*}{\linewidth}{@{\extracolsep{\fill}}clcccccc@{}}
\toprule
\multirow{2}{*}{Training Set} & \multirow{2}{*}{Method} & \multicolumn{2}{c}{Market-1501} & \multicolumn{2}{c}{MSMT17} & \multicolumn{2}{c}{CUHK03} \\
\cmidrule(lr){3-4} \cmidrule(lr){5-6} \cmidrule(lr){7-8}
& & R1 & mAP & R1 & mAP & R1 & mAP \\
\midrule
\multirow{21}{*}{Single Scene} & PCB~\cite{sun2018beyond} & 93.8 & 81.6 & 68.2 & 40.4 & 75.3 & - \\
& ViT-B~\cite{dosovitskiy2020image} & 94.0 & 87.6 & 82.8 & 63.6 & - & - \\
& TransReid~\cite{he2021transreid} & 95.0 & 88.8 & 84.6 & 66.6 & - & - \\
& MGN~\cite{wang2018learning} & 95.1 & 87.5 & 85.1 & 63.7 & - & - \\
& ProFD~\cite{cui2024profd} & 95.1 & 90.0 & - & - & - & - \\
& TransReid*~\cite{he2021transreid} & 95.2 & 89.5 & 86.2 & 69.4 & - & - \\
& AAformer~\cite{zhu2023aaformer} & 95.4 & 87.7 & 83.6 & 63.2 & - & - \\
& AGW~\cite{ye2021deep} & 95.5 & 89.5 & 81.2 & 59.7 & 87.3 & 88.4 \\
& FlipReID~\cite{ni2021flipreid} & 95.5 & 89.6 & 85.6 & 68.0 & - & - \\
& PFD*~\cite{wang2022pose} & 95.5 & 89.7 & 83.8 & 64.4 & - & - \\
& PartFormer~\cite{tan2024partformer} & 95.6 & 89.7 & 85.2 & 68.3 & - & - \\
& LDS~\cite{zang2021learning} & 95.8 & 90.4 & 86.5 & 67.2 & - & - \\
& SCFP~\cite{zhao2024end} & 95.8 & 90.0 & - & - & 84.3 & 82.7 \\
& CPHMNet~\cite{xu2025dual} & 95.9 & 89.9 & 82.5 & 61.6 & - & - \\
& MPN~\cite{ding2020multi} & 96.4 & 90.1 & 83.5 & 62.7 & - & - \\
& CAD-Net~\cite{li2019recover} & - & - & - & - & 82.1 & - \\
& OSNet~\cite{zhou2019omni} & - & - & - & - & 83.8 & 85.4 \\
& ABD-Net~\cite{chen2019abd} & - & - & - & - & 84.3 & 85.9 \\
& INTACT~\cite{cheng2020inter} & - & - & - & - & 86.4 & - \\
& JBIM~\cite{zheng2022joint} & - & - & - & - & 88.7 & 90.3 \\
& PS-HRNet~\cite{zhang2021deep} & - & - & - & - & 92.6 & - \\
\midrule
\multirow{8}{*}{Multiple Scenes} & AIM(R50)~\cite{yang2023good} & 91.9 & 81.6 & 72.6 & 46.4 & 89.3 & 89.9 \\
& ETNDNET(ViT-B)~\cite{dong2023erasing} & 92.7 & 81.6 & 63.8 & 41.5 & 90.8 & 89.7 \\
& PCB~\cite{sun2018beyond} & 93.3 & 83.1 & 69.9 & 48.5 & 87.3 & 90.4 \\
& ETNDNET(R50)~\cite{dong2023erasing} & 93.7 & 84.3 & 75.7 & 50.5 & 90.2 & 89.8 \\
& PFD~\cite{wang2022pose} & 94.3 & 86.8 & 82.0 & 63.3 & 94.1 & 95.5 \\
& AIM(ViT-B)~\cite{yang2023good} & 94.4 & 88.5 & 83.5 & 65.3 & 92.5 & 91.9 \\
& TransReid~\cite{he2021transreid} & 95.0 & 89.7 & 85.8 & 69.4 & 92.9 & 93.6 \\
& ReIDCaps~\cite{huang2019beyond} & 95.6 & 90.6 & 86.6 & 69.6 & 93.3 & 94.6 \\
\midrule
\multirow{2}{*}{Universal} & Baseline (Pure 2D) & 96.5 & 92.7 & \textbf{87.5} & 71.3 & \textbf{96.8} & \textbf{95.9} \\
& Ours & \textbf{96.6} & \textbf{92.9} & 87.4 & \textbf{71.6} & 96.6 & 95.6 \\
\bottomrule
\end{tabular*}
\endgroup
\end{table}

We benchmark our dynamically gated model against a controlled pure-2D baseline with a shared backbone and training protocol. Results in \Cref{tab:holistic,tab:scenes_combined,tab:ablation_fusion,tab:ablation_modality,tab:fusion_evolution} show clearer gains in structurally challenging scenarios while maintaining competitive performance on clean benchmarks. This comparison also clarifies, in this universal setting and practice, when geometry contributes complementary evidence rather than redundant cues.

\begin{table}[!tb]
\centering
\caption{\textbf{Comparison across four ReID scenes.} (a) Clothing-change (PRCC, Celeb-ReID); (b) Occlusion (Occluded-Duke) and Cross-modality (SYSU-MM01). Each subtable uses a method column, and ``-'' denotes results not reported in the papers.}
\label{tab:scenes_combined}
\begingroup
\setlength{\tabcolsep}{1.8pt}
\renewcommand{\arraystretch}{1.03}
\scriptsize

\begin{minipage}[t]{0.48\linewidth}
\centering
\caption*{\scriptsize (a) PRCC and Celeb-ReID}
\setlength{\tabcolsep}{1.0pt}
\begin{tabularx}{\linewidth}{@{}>{\raggedright\arraybackslash}Xcccc@{}}
\toprule
Method & \multicolumn{2}{c}{PRCC} & \multicolumn{2}{c}{Celeb} \\
\cmidrule(lr){2-3} \cmidrule(lr){4-5}
& R1 & mAP & R1 & mAP \\
\midrule
\multicolumn{5}{@{}l}{\textit{Single Scene}} \\
RCSANet~\cite{huang2021clothing} & 31.6 & 31.5 & 55.6 & 11.9 \\
MGN~\cite{wang2018learning} & 33.8 & 35.9 & 49.0 & 10.8 \\
CASE-Net$\ddagger$~\cite{li2021learning} & 39.5 & - & \textbf{66.4} & 18.2 \\
PCB~\cite{sun2018beyond} & 41.8 & 38.7 & 37.1 & 8.2 \\
AFD-Net~\cite{xu2021adversarial} & 42.8 & - & 52.1 & 10.6 \\
ViT-B~\cite{dosovitskiy2020image} & 44.3 & 57.5 & 59.7 & 13.4 \\
TransReid~\cite{he2021transreid} & 45.0 & 57.9 & 58.9 & 14.6 \\
RCSANet$\ddagger$~\cite{huang2021clothing} & 50.2 & 48.6 & 65.3 & 17.5 \\
LightMBN~\cite{herzog2021lightweight} & 50.5 & 51.2 & 57.3 & 14.3 \\
FSAM$\ddagger$~\cite{hong2021fine} & 54.5 & - & - & - \\
IRANet$\ddagger$~\cite{shi2022iranet} & 54.9 & 53.0 & 64.1 & 19.0 \\
CAL~\cite{gu2022clothes} & 55.2 & 55.8 & - & - \\
CLIP3DReID~\cite{liu2024distilling} & \textbf{60.6} & 59.3 & 63.1 & \textbf{19.2} \\
ReIDCaps~\cite{huang2019beyond} & - & - & 51.2 & 9.8 \\
\midrule
\multicolumn{5}{@{}l}{\textit{Multiple Scenes}} \\
PCB~\cite{sun2018beyond} & 44.8 & 57.5 & 10.1 & 1.7 \\
ETNDNET(R50)~\cite{dong2023erasing} & 45.5 & 57.2 & 55.8 & 12.7 \\
ETNDNET(ViT-B)~\cite{dong2023erasing} & 47.6 & 59.2 & 49.8 & 8.5 \\
AIM(ViT-B)~\cite{yang2023good} & 48.2 & 60.7 & 51.3 & 10.5 \\
TransReid~\cite{he2021transreid} & 48.5 & 62.2 & 57.4 & 14.8 \\
ReIDCaps~\cite{huang2019beyond} & 50.3 & 64.4 & 59.9 & 17.7 \\
AIM(R50)~\cite{yang2023good} & 55.4 & 66.0 & 54.5 & 12.4 \\
PFD~\cite{wang2022pose} & 55.5 & 65.8 & 54.0 & 11.0 \\
\midrule
\multicolumn{5}{@{}l}{\textit{Universal}} \\
Baseline (Pure 2D) & 56.0 & 68.0 & 60.0 & 16.5 \\
Ours & 59.0 & \textbf{70.5} & 60.4 & 16.8 \\
\bottomrule
\end{tabularx}
\end{minipage}
\hfill
\begin{minipage}[t]{0.48\linewidth}
\centering
\caption*{\scriptsize (b) Occluded-Duke and SYSU-MM01}
\setlength{\tabcolsep}{0.8pt}
\begin{tabularx}{\linewidth}{@{}>{\raggedright\arraybackslash}Xcccc@{}}
\toprule
Method & \multicolumn{2}{c}{Occ.-Duke} & \multicolumn{2}{c}{SYSU} \\
\cmidrule(lr){2-3} \cmidrule(lr){4-5}
& R1 & mAP & R1 & mAP \\
\midrule
\multicolumn{5}{@{}l}{\textit{Single Scene}} \\
TransReid*~\cite{he2021transreid} & 66.4 & 59.2 & - & - \\
FED~\cite{wang2022feature} & 68.1 & 56.4 & - & - \\
CPHMNet~\cite{xu2025dual} & 68.7 & 61.0 & - & - \\
PFD*~\cite{wang2022pose} & 69.5 & 61.8 & - & - \\
ProFD~\cite{cui2024profd} & 70.8 & 62.8 & - & - \\
SCFP~\cite{zhao2024end} & 70.9 & 63.0 & - & - \\
SCING~\cite{xie2025scing} & 71.1 & 63.4 & - & - \\
PAB-ReID~\cite{chen2024part} & 72.6 & 63.5 & - & - \\
MA-MSA~\cite{jiang2025multi} & 73.3 & 62.9 & - & - \\
HAT~\cite{ye2020visible} & - & - & 55.3 & 53.9 \\
HC~\cite{zhu2020hetero} & - & - & 57.0 & 55.0 \\
VT-ReID~\cite{liu2020parameter} & - & - & 61.7 & 57.5 \\
CM-NAS~\cite{fu2021cm} & - & - & 62.0 & 60.0 \\
\midrule
\multicolumn{5}{@{}l}{\textit{Multiple Scenes}} \\
ETNDNET(ViT-B)~\cite{dong2023erasing} & 44.8 & 38.4 & 51.4 & 43.4 \\
PCB~\cite{sun2018beyond} & 46.7 & 39.0 & 52.7 & 48.3 \\
AIM(R50)~\cite{yang2023good} & 52.3 & 42.9 & 58.2 & 56.4 \\
ETNDNET(R50)~\cite{dong2023erasing} & 58.2 & 48.7 & 55.1 & 50.0 \\
PFD~\cite{wang2022pose} & 59.5 & 52.0 & 62.7 & 62.0 \\
AIM(ViT-B)~\cite{yang2023good} & 62.9 & 54.2 & 60.6 & 60.2 \\
TransReid~\cite{he2021transreid} & 64.2 & 56.6 & 58.8 & 59.5 \\
ReIDCaps~\cite{huang2019beyond} & 66.7 & 58.8 & 63.2 & 59.9 \\
UFFM+AMC~\cite{che2025enhancing} & 68.9 & 61.9 & - & - \\
\midrule
\multicolumn{5}{@{}l}{\textit{Universal}} \\
Baseline (Pure 2D) & 73.9 & 65.5 & 63.1 & 64.3 \\
Ours & \textbf{74.8} & \textbf{65.8} & \textbf{64.3} & \textbf{65.4} \\
\bottomrule
\end{tabularx}
\end{minipage}

\noindent\scriptsize Note: ``$\ddagger$'' denotes contour-based methods; ``*'' means overlapping patch embedding.
\endgroup
\end{table}

\begin{table}[!tb]
\centering
\caption{\textbf{Performance comparison on UAV-Human and AG-ReID.v2.} A$\rightarrow$A denotes the aerial-to-aerial protocol on UAV-Human. A$\rightarrow$C, A$\rightarrow$W, C$\rightarrow$A, and W$\rightarrow$A are AG-ReID.v2 protocols, where A, C, and W denote aerial, CCTV, and wearable views. ``-'' denotes results not reported in the papers.}
\label{tab:uav_ag_results}
\begingroup
\setlength{\tabcolsep}{2.0pt}
\renewcommand{\arraystretch}{1.03}
\scriptsize
\begin{tabular*}{\linewidth}{@{\extracolsep{\fill}}lcccccccccc@{}}
\toprule
\multirow{2}{*}{Method} & \multicolumn{2}{c}{A$\rightarrow$A} & \multicolumn{2}{c}{A$\rightarrow$C} & \multicolumn{2}{c}{A$\rightarrow$W} & \multicolumn{2}{c}{C$\rightarrow$A} & \multicolumn{2}{c}{W$\rightarrow$A} \\
\cmidrule(lr){2-3} \cmidrule(lr){4-5} \cmidrule(lr){6-7} \cmidrule(lr){8-9} \cmidrule(lr){10-11}
& R1 & mAP & R1 & mAP & R1 & mAP & R1 & mAP & R1 & mAP \\
\midrule
AG-ReIDv2~\cite{nguyen2024ag} & \textbf{72.75} & 71.47 & 88.77 & 80.72 & 93.62 & 84.85 & 87.86 & 78.51 & 88.61 & 80.11 \\
SeCap~\cite{wang2025secap} & - & - & 88.12 & 80.84 & 91.44 & 84.01 & 88.24 & 79.99 & 87.56 & 80.15 \\
GSAlign~\cite{li2025gsalign} & - & - & 87.86 & 81.38 & 90.63 & 83.98 & 88.02 & 81.05 & 87.31 & 80.90 \\
\midrule
Baseline (Pure 2D) & 71.4 & \textbf{73.2} & 92.3 & 88.1 & \textbf{94.0} & \textbf{90.2} & \textbf{89.5} & \textbf{83.9} & \textbf{90.7} & \textbf{85.7} \\
Ours & 71.4 & 73.1 & \textbf{92.6} & \textbf{88.8} & 93.7 & 90.1 & 88.9 & \textbf{83.9} & 90.5 & 85.6 \\
\bottomrule
\end{tabular*}
\endgroup
\end{table}

\textbf{Prevention of Negative Transfer on Clean Domains.}
As shown in \Cref{tab:holistic}, on clean domains where texture cues are reliable, our method remains close to the baseline: Market-1501 (96.6\% vs. 96.5\%), MSMT17 (87.4\% vs. 87.5\% R1, 71.6\% vs. 71.3\% mAP), and CUHK03 (96.6\% vs. 96.8\%). This matches our conditional-use hypothesis: in holistic settings, many queries are already resolved by appearance cues, leaving limited room for additional geometry. MSMT17 still gains +0.3\% mAP, while the slight CUHK03 drop shows that geometry is not an always-on benefit. This restrained behavior is consistent with the low gate activation observed in standard holistic scenarios, where the model largely falls back to visual evidence. Instead, the gate mainly helps when appearance becomes ambiguous while avoiding severe negative transfer on clean domains, which is important for universal deployment in practice.

\textbf{Comparison with State-of-the-Art Methods.}
Our method stays competitive across scenarios. On clothing-change (PRCC: 59.0\% R1, +3.0\%; Celeb-ReID: 60.4\% R1, +0.4\%), occlusion (Occluded-Duke: 74.8\% R1, +0.9\%), and cross-modality (SYSU-MM01: 64.3\% R1, +1.2\%), it improves the settings where appearance cues are less reliable. In UAV and wild scenarios, it matches the baseline on UAV-Human Rank-1 (71.4\%), improves AG-ReID.v2 A$\rightarrow$C by +0.3\% Rank-1, and remains competitive on the other transfer protocols.

\subsection{Extensive Ablation Studies}

To isolate more precisely where the gains come from, we ablate both fusion strategy and structural input composition.

\begin{table}[!tb]
\centering
\caption{Ablation Study 1: Multi-Modal Fusion Strategy Analysis on core benchmarks. Rigid 3D fusion actively restricts performance, while Dynamic Gating successfully bridges the modality gap.}
\label{tab:ablation_fusion}
\begingroup
\setlength{\tabcolsep}{1.2pt}
\renewcommand{\arraystretch}{1.03}
\scriptsize
\begin{tabular*}{\linewidth}{@{\extracolsep{\fill}}lcccccccccccccc@{}}
\toprule
\multirow{2}{*}{Method} & \multicolumn{2}{c}{Market} & \multicolumn{2}{c}{MSMT} & \multicolumn{2}{c}{CUHK} & \multicolumn{2}{c}{PRCC} & \multicolumn{2}{c}{Celeb} & \multicolumn{2}{c}{Occ.-Duke} & \multicolumn{2}{c}{SYSU} \\
\cmidrule(lr){2-3} \cmidrule(lr){4-5} \cmidrule(lr){6-7} \cmidrule(lr){8-9} \cmidrule(lr){10-11} \cmidrule(lr){12-13} \cmidrule(lr){14-15}
& R1 & mAP & R1 & mAP & R1 & mAP & R1 & mAP & R1 & mAP & R1 & mAP & R1 & mAP \\
\midrule
Baseline (Pure 2D) & 96.5 & 92.7 & \textbf{87.5} & 71.3 & \textbf{96.8} & \textbf{95.9} & 56.0 & 68.0 & 60.0 & 16.5 & 73.9 & 65.5 & 63.1 & 64.3 \\
+ Naive 3D & \textbf{96.6} & \textbf{93.0} & 87.1 & 71.5 & 96.2 & 95.6 & 55.0 & 67.1 & 59.9 & 16.3 & 73.7 & 65.4 & \textbf{64.6} & \textbf{65.4} \\
+ RG-3D (Ours) & \textbf{96.6} & 92.9 & 87.4 & \textbf{71.6} & 96.6 & 95.6 & \textbf{59.0} & \textbf{70.5} & \textbf{60.4} & \textbf{16.8} & \textbf{74.8} & \textbf{65.8} & 64.3 & \textbf{65.4} \\
\bottomrule
\end{tabular*}
\endgroup
\end{table}

\textbf{Analysis of the Gating Mechanism.}
Our ablation in \Cref{tab:ablation_fusion} verifies the necessity of the Reliability-Aware Gate. Unfiltered structural fusion (Naive 3D Concat) fails to provide stable gains, slightly improving some clean-domain mAP values while regressing on several structurally challenging settings. Relative to the Pure 2D baseline, our dynamically learned reliability gate improves PRCC by +3.0\% Rank-1, Occluded-Duke by +0.9\%, and SYSU-MM01 by +1.2\%, while remaining comparable on standard benchmarks. Its value is especially evident in stability: across the seven reported benchmarks, our gated fusion never underperforms the baseline by more than 0.3\% on either metric, whereas Naive Concat shows larger variance (e.g., PRCC Rank-1 drops 1.0\%). This contrast shows that the key benefit is not simply adding structural capacity, but controlling when geometric evidence participates in the final representation under heterogeneous deployment conditions, thereby avoiding catastrophic negative transfer. Detailed gate activation analysis is provided in \cref{subsec:analysis}.

\begin{table}[!tb]
\centering
\caption{Ablation Study 2: Decoupled SMPL parametric modeling.}
\label{tab:ablation_modality}
\begingroup
\setlength{\tabcolsep}{0.9pt}
\renewcommand{\arraystretch}{1.03}
\scriptsize
\begin{tabularx}{\linewidth}{@{}>{\raggedright\arraybackslash}X*{14}{c}@{}}
\toprule
Method & \multicolumn{2}{c}{Market} & \multicolumn{2}{c}{MSMT} & \multicolumn{2}{c}{CUHK} & \multicolumn{2}{c}{PRCC} & \multicolumn{2}{c}{Celeb} & \multicolumn{2}{c}{Occ.-Duke} & \multicolumn{2}{c}{SYSU} \\
\cmidrule(lr){2-3} \cmidrule(lr){4-5} \cmidrule(lr){6-7} \cmidrule(lr){8-9} \cmidrule(lr){10-11} \cmidrule(lr){12-13} \cmidrule(lr){14-15}
& R1 & mAP & R1 & mAP & R1 & mAP & R1 & mAP & R1 & mAP & R1 & mAP & R1 & mAP \\
\midrule
Black-box SMPL$^\dagger$ & 95.7 & 92.0 & 85.4 & 69.2 & 96.2 & 95.0 & 51.3 & 64.5 & 59.5 & 16.7 & 66.7 & 59.1 & 63.2 & \textbf{65.8} \\
Baseline (Pure 2D) & 96.5 & 92.7 & 87.5 & 71.3 & \textbf{96.8} & \textbf{95.9} & 56.0 & 68.0 & 60.0 & 16.5 & 73.9 & 65.5 & 63.1 & 64.3 \\
Global Params & 96.6 & 92.9 & \textbf{87.6} & 71.5 & 96.7 & 95.7 & 57.2 & 69.0 & 60.2 & \textbf{16.9} & 73.6 & 65.5 & 64.8 & 65.3 \\
Full Topology$^*$ & 96.6 & 92.9 & 87.4 & \textbf{71.6} & 96.6 & 95.6 & \textbf{59.0} & \textbf{70.5} & \textbf{60.4} & 16.8 & \textbf{74.8} & \textbf{65.8} & 64.3 & 65.4 \\
Local Joints & \textbf{96.8} & \textbf{93.0} & 87.5 & \textbf{71.6} & 96.6 & 95.7 & 56.9 & 68.8 & 60.1 & 16.7 & 73.4 & 65.2 & \textbf{65.0} & 65.4 \\
\bottomrule
\multicolumn{15}{@{}p{\linewidth}@{}}{\scriptsize Notation: Black-box=$\mathbf{s}$, Global=$\mathbf{s}_{global}$, Local=$\mathbf{s}_{joint}$, Full=$\mathbf{s}_{global}+\mathbf{s}_{joint}$.} \\
\multicolumn{15}{@{}p{\linewidth}@{}}{\scriptsize $^\dagger$ epoch 120; others epoch 180. \quad $^*$ joint-token pooling only.} \\
\end{tabularx}
\endgroup
\end{table}

\textbf{Necessity of Kinematic-Aware Decoupling.}
Our comparison in \Cref{tab:ablation_modality} isolates the impact of different SMPL modeling strategies. Directly processing the un-decoupled 82-D SMPL vector (Black-box SMPL) induces severe instability characterized by gradient explosion and loss divergence, requiring early termination at epoch 120. This validates our hypothesis: neglecting explicit kinematic parameter semantics entangles invariant factors with volatile variables, destabilizing training. In contrast, explicitly isolating global parameters ($\mathbf{s}_{global}$) and local joint streams ($\mathbf{s}_{joint}$) preserves spatial semantics. Our full topology variant provides the best overall cross-scenario trade-off, while SYSU-MM01 remains a notable exception where local joints achieve slightly higher Rank-1. This suggests that semantically constrained modeling is essential for effective 3D injection, but cross-modality settings can favor the more sensor-stable local-joint representation. This also clarifies the role of the global token. Rather than serving as a direct identity descriptor, it acts as contextual support for joint refinement, allowing global body cues to inform local structure while reducing the risk of view- or sensor-dependent bias in the final representation.

\begin{table}[!tb]
\centering
\caption{Fusion Design Evolution: Comparison against stronger 3D fusion alternatives. Averages are computed over 12 protocols, namely the seven benchmark subset shown in Tables~\ref{tab:holistic}--\ref{tab:ablation_modality}, plus UAV-Human (A$\rightarrow$A) and the four AG-ReID.v2 transfer protocols. The scalar reliability gate is the only 3D-augmented variant that surpasses the pure-2D baseline, achieving Pareto-optimal performance-complexity trade-off.}
\label{tab:fusion_evolution}
\begingroup
\setlength{\tabcolsep}{4.0pt}
\renewcommand{\arraystretch}{1.03}
\scriptsize
\begin{tabular*}{\linewidth}{@{\extracolsep{\fill}}l c c c c@{}}
\toprule
Method & Avg mAP & Avg Rank-1 & Extra Params & Seq. Cost \\
\midrule
HyperNet Prompt$^\dagger$ & 72.1 & 78.5 & $\sim$0.8M & $+K$ tokens \\
3D-SIE (Token Injection) & 72.7 & 76.2 & $\sim$1.2M & $+$24 tokens \\
MoE-Expert (4-Expert Routing) & 73.2 & 77.1 & $\sim$10M & none \\
Pure 2D Baseline & 74.6 & 81.0 & none & none \\
\midrule
Reliability-Gated (Ours) & \textbf{75.0} & \textbf{81.4} & $\sim$0.3M & none \\
\bottomrule
\multicolumn{5}{@{}l@{}}{\scriptsize $^\dagger$ HyperNet Prompt evaluated at epoch 60 due to training instability in subsequent epochs.} \\
\end{tabular*}
\endgroup
\end{table}

\textbf{Fusion Design Evolution.}
To validate the scalar gate, we compared it with three stronger 3D fusion alternatives: MoE-Expert, 3D-SIE, and HyperNet Prompt. Results in \cref{tab:fusion_evolution} show that all three underperform the pure-2D baseline on the 12-protocol average: MoE-Expert drops 1.4\% mAP and 3.9\% Rank-1, 3D-SIE drops 1.9\% mAP and 4.8\% Rank-1, and HyperNet Prompt is unstable, reaching only 72.1\% avg mAP at epoch 60. In contrast, Reliability-Gated is the only 3D-augmented variant that improves over the baseline ($+$0.4\% avg mAP, $+$0.4\% avg Rank-1) with only $\sim$0.3M extra parameters. These stronger alternatives inject geometry too early or too aggressively, forcing noisy structural cues to interact with visual tokens before reliability is assessed. In universal training, where monocular 3D quality varies sharply across datasets and samples, our design instead delays geometric intervention to the final representation stage and modulates structural contribution through cross-modal consistency. The 12-protocol average covers CUHK03, Market-1501, MSMT17, Celeb-ReID, PRCC, Occluded-Duke, SYSU-MM01, UAV-Human (A$\rightarrow$A), and all four AG-ReID.v2 transfer protocols used here for evaluation.

\subsection{Analysis and Visualizations}
\label{subsec:analysis}

\begin{figure}[!tb]
\centering
\begin{minipage}[c]{0.48\linewidth}
\centering
\includegraphics[width=\linewidth]{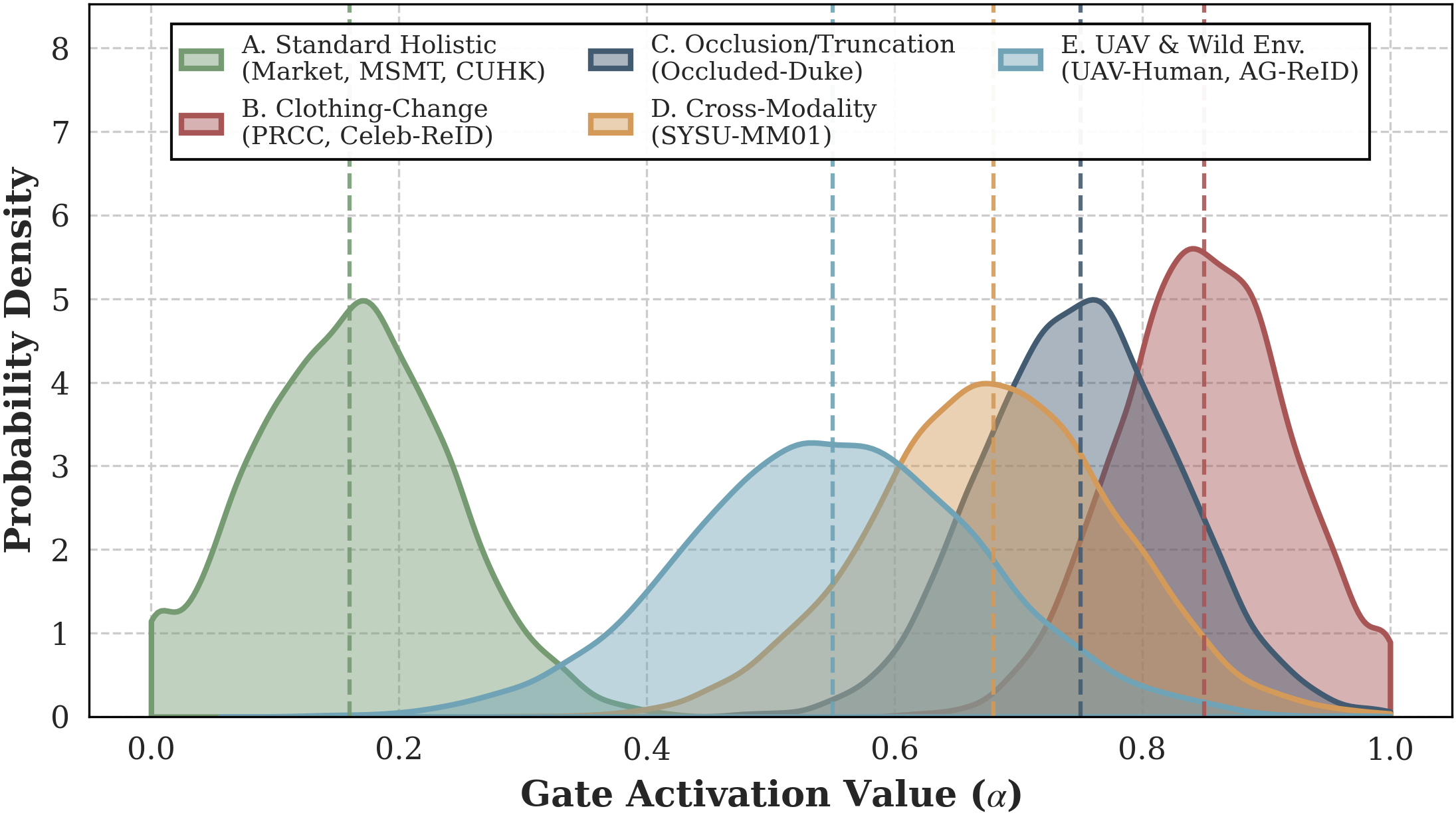}
\end{minipage}
\hfill
\begin{minipage}[c]{0.48\linewidth}
\centering
\includegraphics[width=\linewidth]{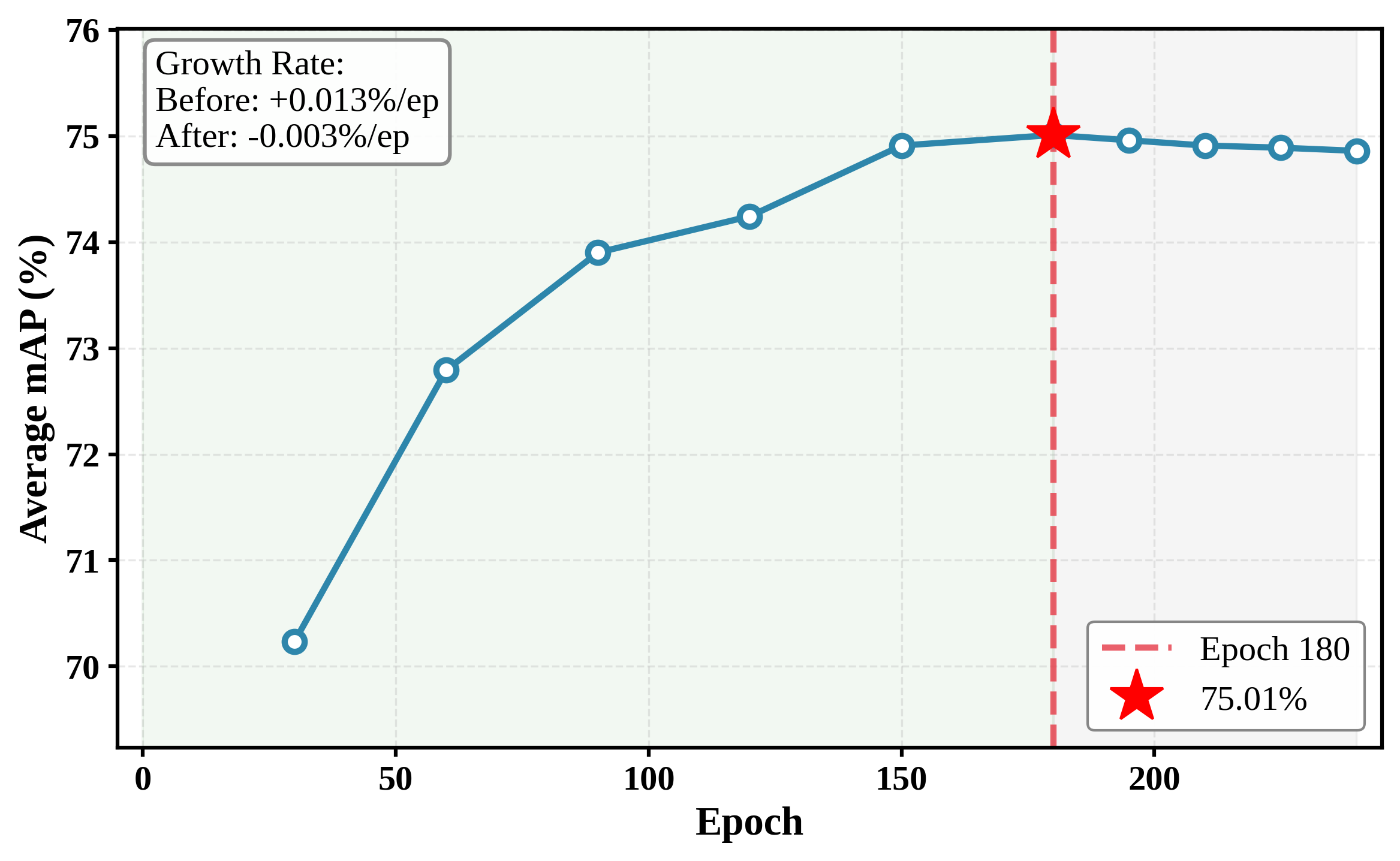}
\end{minipage}

\begin{minipage}[t]{0.48\linewidth}
\caption{\textbf{Dynamic Distribution of Reliability Gate ($\alpha$)} across different representative ReID scenario groups.}
\label{fig:alpha_dist}
\end{minipage}
\hfill
\begin{minipage}[t]{0.48\linewidth}
\caption{\textbf{Training Convergence Analysis} showing average mAP trends over epochs during universal joint training.}
\label{fig:training_convergence}
\end{minipage}
\end{figure}

\textbf{Quantitative Validation of Gate Activation ($\alpha$).}
To better understand gate behavior, we analyze the activation distribution of the scalar $\alpha$ over all query samples at test time. Activation patterns in \cref{fig:alpha_dist} vary clearly across scenario groups. For standard holistic scenarios (Group A: Market-1501, MSMT17, CUHK03), the mean $\alpha$ is approximately $0.15 \pm 0.05$, down-weighting 3D cues and relying on 2D textures. Conversely, for clothing-change scenarios (Group B: PRCC and Celeb-ReID), the mean $\alpha$ increases to $0.82 \pm 0.10$, indicating stronger reliance on geometric posture when appearance cues are unreliable. Occlusion scenarios (Group C: Occluded-Duke, $\mu=0.75$) and cross-modality scenarios (Group D: SYSU-MM01, $\mu=0.68$) also show elevated 3D injection, while UAV and wild scenarios (Group E: UAV-Human and AG-ReID.v2, $\mu=0.55$) remain intermediate, suggesting a more balanced fusion regime under large viewpoint variation. Overall, the learned gate adaptively weights geometric priors by visual-feature reliability across challenging conditions. It acquires this behavior through end-to-end supervision alone, and future work can strengthen the mechanism further by adding explicit geometric quality proxies (\eg 2D keypoint confidence or SMPL fitting scores) as auxiliary reliability signals.

\textbf{Training Convergence and Stability.}
The training curve in \Cref{fig:training_convergence} shows rapid ascent followed by steady convergence, supporting the stability of the proposed reliability-gated optimization. The smooth plateau suggests that scalar reliability modulation avoids the volatility often introduced by more entangled multimodal coupling. This behavior matches our late residual design, which limits cross-stream interference and aligns with the ablations where stronger early-fusion variants underperform or are harder to optimize. Notably, no late-stage oscillation appears after the main performance rise, indicating that the gate settles into a stable cross-modal weighting regime rather than repeatedly over-correcting noisy geometry. This trend is consistent with the clean-domain fallback behavior observed elsewhere, where visual features remain the dominant anchor unless structural cues provide reliably complementary evidence.

\section{Conclusion}

In this paper, we integrated monocular 3D geometric priors into Universal person re-identification through a Reliability-Aware Geometric Injection framework. By decoupling geometric extraction from dynamic utilization, our method explicitly models kinematic joint topology and structure. Coupled with a consistency-aware gating mechanism, the proposed dual-stream residual fusion adaptively leverages geometric cues while maintaining competitive performance on standard holistic benchmarks across diverse deployment scenarios.

Extensive experiments validate this conditional-use perspective, with clear gains on PRCC (+3.0\% Rank-1), Occluded-Duke (+0.9\%), and SYSU-MM01 (+1.2\%), while preserving strong results on standard holistic settings. These results suggest that geometric priors are most valuable when appearance evidence is degraded, and that reliability-aware fallback is essential for avoiding negative transfer in universal deployment. More broadly, our findings indicate that auxiliary geometry should be treated as conditional structural evidence rather than mandatory parallel input, especially in heterogeneous real-world environments where 3D quality can vary substantially across domains and samples.

This interpretation helps reconcile why geometry is highly effective for occlusion, clothing-change, and cross-modality settings, yet should remain selectively suppressed in already well-resolved holistic cases. It also reinforces a broader practical lesson for universal deployment: auxiliary modalities should contribute conditionally, in proportion to their reliability, rather than being fused as uniformly trustworthy signals in practice. Future work will correlate the learned gate with explicit geometric quality metrics and explore temporal geometric aggregation for video-based ReID.

\section*{Acknowledgements}
This work was supported by the National Natural Science Foundation of China under Grant No.~T2541022 and the China Postdoctoral Science Foundation - Hubei Joint Support Program under Grant No.~2025T053HB.

%
%
\bibliographystyle{styles/splncs04}
\bibliography{bib/references}
\end{document}